\pdfoutput=1
\documentclass[11pt]{article}

\usepackage[final]{acl}

\usepackage{times}
\usepackage{latexsym}
 
\usepackage[T1]{fontenc}
\usepackage[utf8]{inputenc}

\usepackage{microtype}

\usepackage{inconsolata}

\usepackage{graphicx}
\usepackage{multirow}
\usepackage{amsmath}
\usepackage{amssymb}
\usepackage{booktabs}
\usepackage{subcaption}
\usepackage{bm}
\usepackage{tcolorbox}
\usepackage{enumitem}

\usepackage{cleveref}
\usepackage{makecell}
\usepackage{booktabs}

\usepackage{color}
\usepackage{xspace}
\newcommand{\wavedetect}{\textsc{WaveDetect}\xspace}
\newcommand{\Ours}{\textsc{WaveDetect}\xspace}
\DeclareRobustCommand{\githublogo}{\raisebox{-0.15ex}{\includegraphics[height=3.5ex]{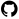}}}
\DeclareRobustCommand{\huggingfacelogo}{\raisebox{-0.15ex}{\includegraphics[height=3.5ex]{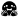}}}

\title{\textsc{WaveDetect}: Robust Framework for Machine-Generated Text Detection via Wavelet Transform}

\author{
 \textbf{Zhichen Liu\textsuperscript{*}},
 \textbf{Kaitong Qin\textsuperscript{*}},
 \textbf{Linhan He\textsuperscript{}},
 \textbf{Yang Xu\textsuperscript{\textdagger}}, % correspondence
\\
 Dept. of Computer Science and Engineering, Southern University of Science and Technology
\\
 \small{
 \textsuperscript{*}\textbf{Equal contribution}, \quad
 \textsuperscript{\textdagger}\textbf{Correspondence:} \href{mailto:xuyang@sustech.edu.cn}{xuyang@sustech.edu.cn}
}
\\
\small{
 \href{https://github.com/KaitongQin/WaveDetect}{\githublogo}
 \quad
 \href{https://huggingface.co/KaitongQin/WaveDetect}{\huggingfacelogo}
}
}

\begin{document}
\maketitle
\begin{abstract}
As Large Language Models asymptotically approach human-level fluency in natural language generation, solely relying on surface-level semantic artifacts for detecting LLM-generated texts has become increasingly precarious. Existing detectors often falter when facing three critical challenges: adversarial perturbations, cross-domain shifts, and the rapid temporal evolution of the foundation model. To address these issues, we propose \wavedetect, a novel framework that reformulates text detection as a signal processing task within the time-frequency domain. Unlike previous methods that analyze static token probability distributions, \wavedetect models the generated output as a probability signal, upon which a differentiable Continuous Wavelet Transform is applied to convert them into learnable spectral representations. This process reveals the intrinsic ``spectral fingerprints'' in machine-generated texts--patterns that remain invisible in time domain. Comprehensive evaluations on three well-curated datasets (RAID, EvoBench, and Domain-Shift) show that our method achieves a new state-of-the-art. It not only achieves superior accuracy but also exhibits remarkable robustness against sophisticated attacks, generalization across out-of-distribution topics and unseen evolving LLMs. Our results validate the efficacy of spectral analysis as a promising paradigm for LLM-generated texts detection. 
\end{abstract}

\section{Introduction} \label{sec:intro}
The advent of Large Language Models (LLMs) has fundamentally reshaped the landscape of natural language generation, bringing us asymptotically close to a reality where machine-generated text is nearly indistinguishable from human writing. While this capability offers immense productivity gains, it simultaneously precipitates a crisis of information integrity \cite{threat}. The potential for misuse ranges from academic dishonesty and automated plagiarism to the industrial-scale proliferation of disinformation \cite{iclr26}. Consequently, the development of robust Machine-Generated Text (MGT) detection systems has escalated from a technical curiosity to a critical imperative for AI safety.

Despite urgency of the issue, the detection landscape is currently locked in an asymmetric arms race. As LLMs scale in parameter count and reasoning capability, the subtle artifacts traditionally used for detection are rapidly vanishing, such as perplexity gaps or entropy differences \cite{gltr}. Contemporary detection frameworks grapple with three pivotal challenges for deployment:
\begin{itemize}[leftmargin=*, itemsep=0em, topsep=0.5em]
    \item \textbf{Robustness against Adversarial Attacks:} Real-world adversaries actively employ obfuscation techniques, such as paraphrasing or synonym substitution, to evade detection. Prior work indicates that semantic-based detectors are notoriously brittle, often collapsing under minimal perturbations \cite{raid, attackstresstest}.
    \item \textbf{Temporal Stability (Model Evolution):} The rapid iteration of LLMs (e.g., from GPT-3 to GPT-4o) renders detectors obsolete effectively overnight. A core challenge is preventing detectors from overfitting to the idiosyncrasies of legacy models, thereby ensuring they remain effective against unseen, next-generation architectures \cite{evobench}.
    \item \textbf{Domain Generalization:} Detectors trained on general corpora frequently fail when applied to specialized fields such as law or medicine \cite{divscore}. A reliable detector must discern the fundamental distributional signature of machine generation, rather than latching onto domain-specific surface features.
\end{itemize}

\begin{figure*}[htbp]
    \centering
    \includegraphics[width=\linewidth]{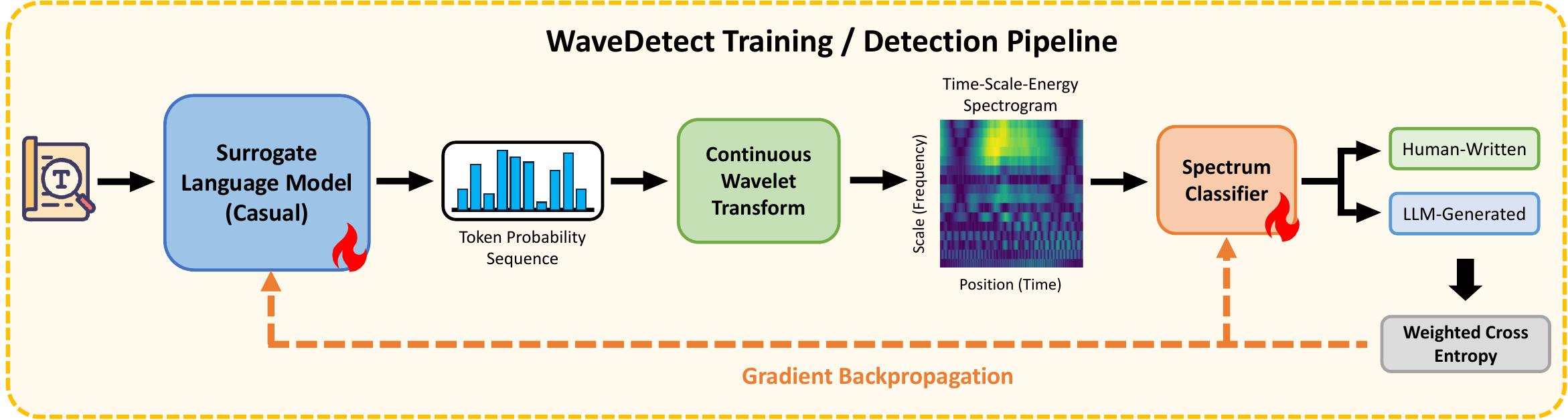}
    \caption{The overall workflow of \wavedetect. A surrogate language model is used to obtain probabilities sequence, CWT transforms the sequential features into time-frequency domain, and a CNN is used to extract spectral features for classification.}
    \label{fig:main}
\end{figure*}

Existing methodologies are typically categorized into training-based and zero-shot approaches \cite{survey}. Zero-shot methods \cite{detectgpt, binoculars}, favored for their interpretability, often lack the resilience required for high-stakes environments such as under attack \cite{attackstresstest}. Conversely, training-based supervised classifiers \cite{openai-roberta} offer higher accuracy but are prone to the aforementioned overfitting, struggling to generalize out-of-distribution domains. We hypothesize that this limitation stems from a reliance on the \textit{time-domain} representation of text (i.e., the raw sequence of tokens), where the boundary between human and machine semantics is increasingly blurred.

To overcome these limitations, we propose a paradigm shift from semantic analysis to \textbf{spectral analysis}. Recent psycholinguistics-inspired studies \cite{fouriergpt, tdt} suggest that while human and machine texts may be semantically close, they exhibit distinct regular patterns in their probability distributions, fingerprint-like features of the MGTs. Nevertheless, simplistic Fourier-based methods (e.g., FourierGPT \cite{fouriergpt}) tend to lose essential temporal localization information, which is pivotal for capturing fine-grained differences.

Therefore, drawing on these insights, we hypothesize that while human-written texts (HWTs) and MGTs may be semantically indistinguishable, they exhibit distinct rhythmic patterns in their probability distributions. 
Based on this assumption, we introduce \textbf{\wavedetect}, a novel framework that reformulates text detection as a signal processing problem in the time-frequency domain. By transforming the sequence of token probabilities into continuous signals and decomposing them via Continuous Wavelet Transform (CWT) \cite{cwt}, \Ours captures both global spectral features and local transient features, allowing a trainable detector to learn the underlying ``spectral fingerprints'' that are invariant to semantic paraphrasing and domain shifts.

We conduct a comprehensive evaluation within RAID \cite{raid}, EvoBench \cite{evobench}, and Domain Shift \cite{divscore}. To summarize, our main contributions are as follows:
\begin{itemize}[leftmargin=*, itemsep=0em, topsep=0.5em]
    \item We propose \textbf{\wavedetect}, a novel MGT detector that leverages CWT to extract robust time-frequency features, effectively characterizing the temporal dynamics of token probabilities beyond surface text features.
    \item New state-of-the-art performance is achieved in detection robustness, significantly outperforming baselines under challenges from adversarial attacks, model evolution, and domain shifts.
    \item Visualizations of the spectrum features show that human and machine text distributions can be disentangled in spectral space.
\end{itemize}

\section{Related Work}

\subsection{Machine-Generated Text Detection}
Early approaches trained BERT-like models for machine text classification, as seen in \citep{openai-roberta, llm-detectaive}. Instead of direct training, \citet{detective} employed multi-level contrastive learning to train the detector. Specifically, RADAR \citep{radar} utilized adversarial training to develop a detector, ensuring robustness against adversarial attacks.
Other zero-shot methods exploit inherent behavioral differences between machine-generated and human-written text. 
For example, DetectGPT \cite{detectgpt} observed that machine-generated text is more sensitive to minor perturbations; substituting words in machine text leads to significant changes in the loss. Fast-DetectGPT \cite{fastdetectgpt} subsequently found that such perturbations can be simulated directly during the final vocabulary decision step, which largely accelerates detection speed. Binoculars \cite{binoculars} normalizes the perplexity of the input text using the inherent differences between two scoring models, establishing a strong zero-shot baseline. RepreGuard \cite{repreguard} uses the internal hidden state representations of a surrogate model to calculate a hyperplane that distinguishes between machine and human text, achieving stable out-of-distribution (OOD) performance.

\subsection{Spectrum Methods}
Previous studies have demonstrated significant differences between the spectra of human-written and model-generated text \cite{face, face2}. FourierGPT \cite{fouriergpt} leverages these differences to detect machine-generated text by applying the Fourier transform to the surprisal sequence of the input text. Based on the same assumption, \citet{tdt} employed the Wavelet transform instead of the Fourier transform, allowing for the utilization of both spectral and temporal information. These works demonstrate the effectiveness of detectors based on spectral analysis, highlighting their high potential for future developments in machine text detection.

\section{Methodology}
In this section, we introduce the workflow of \wavedetect, with focus on the technique of extracting the latent spectral fingerprints of the input text. 
As illustrated in \Cref{fig:main}, we build a signal processing-inspired pipeline, in which discrete input tokens are transformed into continuous-valued probability signals, and then decompose them via a differentiable CWT, and finally leverage a convolutional neural network (CNN) to capture the most salient time-frequency features.

\subsection{Signal Extraction and Spectral Hypothesis}

Formally, let $\mathcal{X} = \{x_1, \dots, x_N\}$ denote a sequence of tokens. We formulate detection as a binary hypothesis test between human distribution $\mathcal{D}_{\mathcal{H}}$ and machine distribution $\mathcal{D}_{\mathcal{M}}$. We define a mapping $\Phi: \mathcal{V}^N \to \mathbb{R}^N$ utilizing a surrogate language model $\mathcal{M}_\theta$ to transform discrete tokens into a continuous probability signal $\mathbf{p}$:

\begin{equation}
\mathbf{p}_t = \Phi(x_t) = P_{\mathcal{M}_\theta}(x_t \mid x{<t}), \quad t=1,\dots,N
\end{equation}

We hypothesize that human and machine signals exhibit distinct time-frequency characteristics due to their disparate generative mechanisms. Human writing is inherently stochastic and contextually diverse, characterized by a high-entropy selection process that introduces irregular ``spikes'' and chaotic, transient fluctuations in $\mathbf{p}_t$. In contrast, machine generation is strictly constrained by decoding strategies such as top-$k$, top-$p$ (nucleus sampling), and repetition penalties:
% \begin{equation}
$x_t \sim \text{Decode}\left( P_{\text{gen}}(\cdot \mid x_{<t}) \right)$
% \end{equation}
These mechanisms mechanically truncate the low-probability tail and manipulate token distributions. This artificially smooths the probability signal and forces it into more structured, periodic rhythmic patterns, thereby fundamentally suppressing the high-frequency transients naturally found in human text. Therefore, detecting $\mathcal{D}_{\mathcal{M}}$ is equivalent to capturing these localized rhythmic and transient disparities.

\subsection{Continuous Wavelet Transform}
Since natural language probability signals are inherently \textit{non-stationary}, as their statistical properties shift with context, global spectral methods like Fourier Transform fail to capture localized transient features. We therefore employ CWT to decompose $\mathbf{p}$ into a time-frequency representation.

We select the Morlet wavelet \cite{morlet} as the mother wavelet $\psi(t)$ due to its Gaussian envelope, which offers an optimal balance between time and frequency resolution:
\begin{equation}
\psi(t) = \pi^{-1/4} e^{i \omega_0 t} e^{-t^2 / 2}
\end{equation}
where $\omega_0$ is the central frequency. The CWT of the signal $\mathbf{p}(t)$ at scale $s$ and translation $\tau$ is defined as the convolution of the signal with the scaled and translated wavelet:
% $$W(s, \tau) = \frac{1}{\sqrt{s}} \int_{-\infty}^{\infty} \mathbf{p}(t) \psi^*\left(\frac{t - \tau}{s}\right) dt$$
% The CWT coefficients $W(s, \tau)$ at scale $s$ and translation $\tau$ are computed as:
\begin{equation}
W(s, \tau) = \frac{1}{\sqrt{s}} \int_{-\infty}^{\infty} \mathbf{p}(t) \psi^*\left(\frac{t - \tau}{s}\right) dt
\end{equation}
In our implementation, we treat this as a fixed-weight 1D convolution layer with a bank of kernels corresponding to distinct scales. The magnitude $|W| \in \mathbb{R}^{K \times N}$ serves as the spectral map. Since there exists meaningless padding tokens in batch training, we apply right-side zero-padding on the spectrum as:
\begin{equation}
\mathbf{S}_{ij} = 
\begin{cases} 
|W(s_i, \tau_j)| & \text{if } j \le N \\
0 & \text{otherwise}
\end{cases}
\end{equation}
which ensures that the spectral features of the valid text are preserved without distortion from padding artifacts in the time domain.

\subsection{Spectrum Map Feature Extraction}
The spectral map $\mathbf{S}$ serves as a visual fingerprint of the text's generation process. To capture the special spectrum patterns between MGT and HWT, we feed $\mathbf{S}$ into a lightweight CNN, effectively treating the detection task as image classification. The network outputs a probability $\hat{y} = \sigma(\text{CNN}(|\mathbf{W}|))$ to classify the text source as human or model.

\subsection{Optimization Strategy} \label{sec:method-optim}
To handle the optimization gap between the pre-trained surrogate model and the randomly initialized CNN, we employ a two-stage training protocol: (1) \textbf{Warm-up:} Freezing $\mathcal{M}_\theta$ to align the CNN with spectral features, it forces the CNN to learn extracting robust features from the intrinsic spectral patterns of the pre-trained surrogate's output; (2) \textbf{Joint Training:} Unfreezing $\mathcal{M}_\theta$ to refine the probability distribution for detection, it allows the surrogate model to dig deeper in the differences between MGT and HWT, and demonstrate such differences in spectrum space. 

We optimize using a Weighted Cross-Entropy (WCE) loss to address class imbalance and maximize the decision margin:
\begin{equation}
\mathcal{L} = - \frac{1}{B} \sum_{i=1}^{B} \left[ \alpha y_i \log(\hat{y}_i) + (1-y_i) \log(1 - \hat{y}_i) \right]
\end{equation}
where $\alpha$ balances the contributions of positive (machine) and negative (human) samples. Compared to contrastive learning based approaches \cite{detective}, WCE is easier to implement with distributed data parallelism, making it more suitable for large-scale training.

\section{Experiments}
We conduct a comprehensive suite of experiments to validate our method, aiming to answer the three research questions mentioned in \Cref{sec:intro}:
\begin{itemize}[leftmargin=*, itemsep=0em, topsep=0.5em]
\item \textbf{RQ1:} Is \wavedetect's performance robust against adversarial attack on in-distribution dataset?
\item \textbf{RQ2:} Can we maintain the performance as LLMs continually evolve?
\item \textbf{RQ3:} Can we generalize the detection capabilities to domains that are unseen before?
\end{itemize}

\subsection{Experiment Setup} \label{sec:experiment setup}

\paragraph{Training Data Construction} To ensure the model learns generalized spectral features rather than overfitting to specific content, we utilize the RAID benchmark \cite{raid} as our primary training corpus. RAID encompasses 11 diverse LLMs employing 3 distinct decoding strategies across 8 domains (details in Appendix \ref{app:raid}). To strictly mitigate data leakage, we implement a prompt-level splitting strategy. We randomly withhold 150 source prompts (100 for testing, 50 for validation) and all their corresponding generated texts. The remaining prompts and their derivations constitute the training set (approx. 6 million samples). This protocol ensures that the detector never encounters generations derived from the same semantic context during inference, forcing it to rely on structural generation artifacts. In this work, we denote the training set as RAID-all, and the subset that excludes all sampling and adversarial augmentation as RAID-base, and the test set as RAID-test\footnote{The formal evaluation on RAID's test set requires a pull-request in RAID's github repository: \url{https://github.com/liamdugan/raid}. We manually conduct train-test split in this study for a convenient local evaluation.}.

\begin{table*}[htbp]
    \centering
    \resizebox{\linewidth}{!}{
        \begin{tabular}{l|c c c c|c c|c|c c|c c|c}
        \toprule
        \multicolumn{13}{c}{\textbf{AUROC} ($\uparrow$)} \\
        \midrule
        Detectors & ChatGPT & GPT-2 & GPT-3 & GPT-4 & Cohere & Cohere-C & Llama-C & Mistral & Mistral-C & MPT & MPT-C & Total  \\
        \midrule
        RoBERTa & 0.6825 & 0.6144 & 0.5985 & 0.5954 & 0.5449 & 0.6529 & 0.6955 & 0.5267 & 0.6921 & 0.4400 & 0.5772 & 0.6018  \\
        RADAR & 0.8968 & 0.7881 & 0.8750 & 0.8661 & 0.7516 & 0.8375 & 0.8777 & 0.7185 & 0.8823 & 0.7589 & 0.8518 & 0.8277  \\
        \midrule
        Fast-DetectGPT & 0.8715 & 0.7735 & 0.8578 & 0.8014 & 0.8060 & 0.8277 & 0.8653 & 0.6513 & 0.8023 & 0.5391 & 0.6369 & 0.7666   \\
        Binoculars & 0.9483 & 0.7803 & 0.9473 & 0.9047 & 0.8813 & 0.9213 & 0.9419 & 0.6962 & 0.9008 & 0.5751 & 0.7582 & 0.8414  \\
        FourierGPT & 0.7454 & 0.8035 & 0.8379 & 0.7854 & 0.6885 & 0.7410 & 0.7855 & 0.7537 & 0.7110 & 0.8416 & 0.7764 & 0.7700  \\ 
        RepreGuard & 0.8184 & 0.5501 & 0.5758 & 0.7844 & 0.5681 & 0.6525 & 0.8538 & 0.5510 & 0.7691 & 0.4395 & 0.6528 & 0.6560  \\ 
        \midrule
        \textsc{WaveDetect-base} & 0.8986 & 0.7548 & 0.9265 & 0.8696 & 0.8415 & 0.8836 & 0.8929 & 0.7313 & 0.8966 & 0.6528 & 0.8460 & 0.8359 \\
        \textsc{WaveDetect-all} & \textbf{0.9944} & \textbf{0.9597} & \textbf{0.9953} & \textbf{0.9924} & \textbf{0.9462} & \textbf{0.9737} & \textbf{0.9957} & \textbf{0.9638} & \textbf{0.9931} & \textbf{0.9601} & \textbf{0.9895} & \textbf{0.9785}  \\
        \toprule
        \multicolumn{13}{c}{\textbf{TPR@0.1\%FPR} ($\uparrow$)} \\
        \midrule
        RoBERTa & 0.0572 & 0.0385 & 0.0111 & 0.0185 & 0.0080 & 0.0614 & 0.0671 & 0.0175 & 0.1240 & 0.0127 & 0.0791 & 0.0450  \\
        RADAR & 0.0060 & 0.0078 & 0.0312 & 0.0025 & 0.0010 & 0.0105 & 0.0029 & 0.0027 & 0.0094 & 0.0079 & 0.0193 & 0.0092  \\
        \midrule
        Fast-DetectGPT & 0.0051 & 0.0045 & 0.0047 & 0.0038 & 0.0036 & 0.0041 & 0.0054 & 0.0031 & 0.0043 & 0.0028 & 0.0026 & 0.0040  \\
        Binoculars & 0.4838 & 0.3094 & 0.5470 & 0.3212 & 0.1667 & 0.3854 & 0.2257 & 0.2256 & 0.4468 & 0.1979 & 0.3284 & 0.3584 \\
        FourierGPT & 0.0101 & 0.0976 & 0.0216 & 0.0096 & 0.0013 & 0.0021 & 0.0123 & 0.0496 & 0.0013 & 0.0859 & 0.0015 & 0.0266  \\
        RepreGuard & 0.0687 & 0.0192 & 0.0145 & 0.0619 & 0.0119 & 0.019 & 0.0971 & 0.013 & 0.0539 & 0.012 & 0.0108 & 0.0347  \\
        \midrule
        \textsc{WaveDetect-base} & 0.0743 & 0.0366&	0.2642	& 0.0283 & 0.0354 & 0.1003 & 0.0392 & 0.0238 & 0.0971 & 0.0135 & 0.1488 & 0.0783 \\
        \textsc{WaveDetect-all} & \textbf{0.7141} & \textbf{0.3753} & \textbf{0.6751} & \textbf{0.5937} & \textbf{0.4466} & \textbf{0.5404} & \textbf{0.6709} & \textbf{0.4054} & \textbf{0.6512} & \textbf{0.3643} & \textbf{0.5803} & \textbf{0.5470} \\
        \bottomrule
        \end{tabular}
    }
    \caption{\textbf{Detection Performance on RAID Test Set.} We report the Area Under the Receiver Operating Characteristic (AUROC) and the True Positive Rate at 0.1\% False Positive Rate (TPR@0.1\%FPR). The best results are highlighted in \textbf{bold}.}
    \label{tab:raid}
\end{table*}

\paragraph{Evaluation Protocol} Echoing the challenges outlined in \Cref{sec:intro}, we evaluate \wavedetect across three rigorous dimensions:
\begin{enumerate}[leftmargin=*, itemsep=0em, topsep=0.5em] 
\item \textbf{Adversarial Robustness (RAID-test):} We utilize the withheld RAID test set to evaluate performance under standard conditions and 11 adversarial attack scenarios (e.g., homoglyph substitution, zero-width injection, and paraphrasing). This assesses the detector's resilience to active evasion attempts.
\item \textbf{Temporal Stationarity (EvoBench):} To assess robustness against "Model Evolution," we employ EvoBench \cite{evobench}, which contains generations from evolving versions of LLMs (e.g., GPT-4o-0513 $\to$ GPT-4o-0806 $\to$ GPT-4o-1120). Crucially, these newer models and the closed-source families (Claude-3.5, Gemini-1.5) were never seen during \wavedetect's training.
\item \textbf{OOD Domain Generalization:} We utilize the dataset from DivScore \cite{divscore} to test Out-of-Distribution (OOD) generalization. The dataset includes two domains: \textit{Medical} (PubMed \cite{pubmedqa}, MIMIC \cite{mimic}) and \textit{Legal} (LawStack \cite{lawstack}, OALC \cite{olac}). These distributions differ significantly from the training set, and we ensure that they were never seen during training.
\end{enumerate}

\paragraph{Baseline Detectors} We benchmark \wavedetect against a comprehensive suite of detectors, categorized by their operational paradigm:

\noindent\textbf{$\bullet$ Supervised Detectors:} We include \textbf{OpenAI-RoBERTa} \cite{openai-roberta}, a classic RoBERTa-large model fine-tuned for binary classification; and \textbf{RADAR} \cite{radar}, a detector trained by an adversarial training framework, where a paraphraser and a detector play a min-max game to improve robustness of text detection.

\noindent\textbf{$\bullet$ Zero-Shot Methods:} We select \textbf{Fast-DetectGPT} \cite{fastdetectgpt}, an efficient successor to DetectGPT, based on the hypothesis that MGT lies in regions of positive curvature within the model's log-probability landscape. It estimates the conditional probability curvature to distinguish machine text from human writing; \textbf{Binoculars} \cite{binoculars}, which calculates a score based on the ratio of the perplexity computed by an observer model to that of a performer model; \textbf{RepreGuard} \cite{repreguard}, which assumes that human and machine texts form distinct manifolds in the latent space. It applies PCA on human and model texts' hidden states to define a separation hyperplane for classification; and \textbf{FourierGPT} \cite{fouriergpt}, a most relevant baseline which applies Fourier transform to the surprisal sequences and conduct classification in the spectrum space. Comparing against this highlights the advantage of our Wavelet-based approach (Time-Frequency vs. Frequency-only).

\paragraph{Implementation Details} \ 

\noindent\textbf{$\bullet$ Architecture:} We use \textbf{Qwen2.5-0.5B-Base} \cite{qwen2.5} as the surrogate probability estimator $\mathcal{M}_{\theta}$ for its efficiency and strong capabilities. The spectral encoder is based on a ResNet-18 backbone modified for single-channel spectrogram input.

\noindent\textbf{$\bullet$ Training Strategy:} We adopt the two-stage protocol described in \Cref{sec:method-optim} using the AdamW optimizer with a batch size of 64 and epochs of 3. In the warm-up stage, the CNN is trained from scratch with a learning rate of $1\times 10^{-3}$, and $\mathcal{M}_{\theta}$ is frozen. In the joint training stage, we jointly train the entire \wavedetect (all parameters) with a learning rate of $1\times 10^{-5}$. 
To address the data imbalance (HWT vs. MGT) issue, we apply the Weighted Cross-Entropy loss with a weight ratio of 9:1 (HWT:MGT). All experiments are conducted on $4 \times$ NVIDIA RTX 6000 Ada GPUs. 
The detector trained on RAID-all is referred to as \textsc{WaveDetect-all}, and the one trained on RAID-base is \textsc{WaveDetect-base}.

\begin{table*}[htbp]
    \centering
    \resizebox{\linewidth}{!}{
    \begin{tabular}{l|c c c|c c c|c c c|c c c}
    \toprule
    \multirow{2}*{Detectors} & \multicolumn{3}{c|}{GPT-4o} & \multicolumn{3}{c|}{GPT-4} & \multicolumn{3}{c|}{Claude-Sonnet} & \multicolumn{3}{c}{Gemini-1.5-Flash}  \\
    % \midrule
        & AUC-m & Std & max$|\Delta|$ & AUC-m & Std & max$|\Delta|$ & AUC-m & Std & max$|\Delta|$ & AUC-m & Std & max$|\Delta|$ \\
    \midrule
    RoBERTa & 0.6216 & 0.0802 & 0.1507 & 0.6866 & 0.0154 & 0.0376 & 0.6866 & 0.0226 & 0.0307 & 0.3132 & 0.2578 & 0.5453 \\
    RADAR & 0.7851 & 0.0153 & 0.0292 & 0.7825 & 0.0122 & 0.0150 & 0.7803 & 0.0309 & 0.0742 & 0.7328 & \textbf{0.0129} & \underline{0.0308} \\
    \midrule
    Fast-DetectGPT & 0.7559 & 0.0253 & 0.0676 & 0.7621 & 0.0120 & 0.0237 & 0.8007 & 0.0490 & 0.0982 & 0.7371 & 0.0557 & 0.1104 \\
    Binoculars & \underline{0.8364} & 0.0118 & 0.0323 & \underline{0.8276} & \underline{0.0036} & 0.0087 & \underline{0.8406} & 0.0334 & 0.0644 & \textbf{0.8013} & 0.0398 & 0.0801 \\
    FourierGPT & 0.6489 & 0.0423 & 0.0882 & 0.5984 & 0.0172 & 0.0306 & 0.6728 & \underline{0.0149} & \underline{0.0264} & 0.5867 & \underline{0.0222} & 0.0381 \\
    RepreGuard & 0.6618 & 0.0117 & \underline{0.0172} & 0.6562 & 0.0081 & 0.0160 & 0.6434 & \textbf{0.0116} & \textbf{0.0231} & 0.6319 & 0.0295 & \textbf{0.0298} \\
    \midrule
    \textsc{WaveDetect-base} & \textbf{0.8536} & \textbf{0.0048} & \textbf{0.0087} & \textbf{0.8404} & \textbf{0.0023} & \textbf{0.0052} & \textbf{0.8767} & 0.0304 & 0.0682 & \underline{0.7844} & 0.0404 & 0.0869 \\
    \textsc{WaveDetect-all} & 0.8145 & \underline{0.0063} & 0.0179 & 0.8009 & 0.0049 & \underline{0.0086} & 0.8180 & 0.0251 & 0.0572 & 0.7390 & 0.0380 & 0.0794 \\
    \bottomrule
    \end{tabular}
    }
    \caption{\textbf{Temporal Robustness on EvoBench.} We evaluate detectors on evolving versions of LLMs. $\text{AUC-m}$ denotes the mean AUROC across versions, $\text{Std}$ is the standard deviation, and max$|\Delta|$ represents the maximum performance changes between versions. Lower Std and max$|\Delta|$ indicate better stability.}
    \label{tab:evobench}
\end{table*}

\subsection{Basic Performance on RAID-test}
Table \ref{tab:raid} summarizes the performance of \wavedetect against four representative baselines across 11 diverse source models. \wavedetect establishes a decisive state-of-the-art performance on RAID-test dataset, achieving an average AUROC of \textbf{0.9785}. This represents a substantial improvement over the strongest zero-shot baseline, Binoculars (0.8414), and the supervised baseline, RADAR (0.8277). Notably, purely semantic-based supervised methods like RoBERTa struggle significantly (0.6018), often degenerating to near-random guessing on unseen models (e.g., MPT: 0.44). 
This supports our hypothesis that spectral features are more invariant across different generator architectures than surface-level semantic artifacts. 

In real-world scenarios, maintaining a low False Positive Rate (FPR) is critical to prevent false accusations. As shown in the bottom half of Table \ref{tab:raid}, \wavedetect demonstrates exceptional reliability in this strict regime. It achieves a total TPR of \textbf{54.70\%} at 0.1\% FPR, outperforming Binoculars by nearly 19 percentage points. In contrast, other methods fail to distinguish machine text effectively at such high precision thresholds. This result highlights \wavedetect's potential for deployment in sensitive applications such as academic integrity review.

\subsection{RQ1: Robustness against Adversarial Attacks}
\begin{figure}[tbp]
    \centering
    \includegraphics[width=\linewidth]{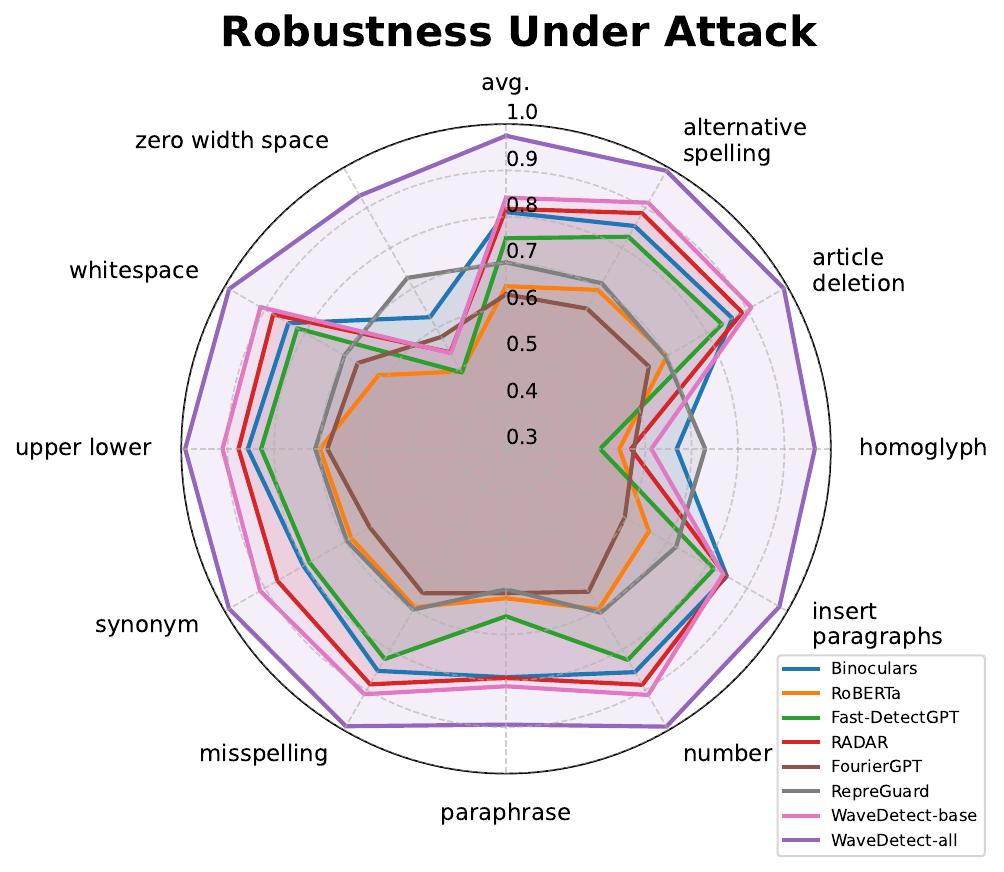}
    \caption{\textbf{Robustness against Adversarial Attacks.} We visualize the detection performance (AUROC) under 11 different attack methods using a radar chart. \wavedetect exhibits a dominant performance envelope, completely enclosing all baseline methods, maintaining a larger consistent envelope compared to baselines.}
    \label{fig:attack}
\end{figure}

Real-world deployment of detectors often faces adversarial attempts to bypass detection. We evaluate \wavedetect against 11 diverse attack methods ranging from character-level noising to semantic rewriting. \Cref{fig:attack} visualizes the performance comparison using a radar chart.

As illustrated in \Cref{fig:attack}, \wavedetect (\textsc{WaveDetect-all}, represented by the purple line) demonstrates a dominant advantage in adversarial attacks, largely outperforming all the baselines. In the hardest two attack types, zero-width space and homoglyph, \wavedetect maintains a very high performance (above 90\% acc.), while most other baselines suffer from significant performance degradation. This indicates that detectors trained on spectral features exhibit greater robustness compared to those based on sequential features in time domain.

\begin{table*}[htbp]
    \centering
    \resizebox{0.8\linewidth}{!}{
    \begin{tabular}{l|c c c|c c c|c}
    \toprule
    \multirow{2}*{Detectors} & \multicolumn{3}{c|}{Medical} & \multicolumn{3}{c|}{Legal} & \multirow{2}*{All Avg.} \\
    % \cline{2-7}
        & MIMIC & PubMed & Avg. & LawStack & OALC & Avg. & \\
    \midrule
    RoBERTa & 0.6078 & 0.3941 & 0.5010 & 0.2142 & 0.4839 & 0.3490 & 0.4250  \\ 
    RADAR & 0.2319 & 0.1852 & 0.2085 & 0.6199 & 0.6264 & 0.6232 & 0.4159  \\ 
    \midrule
    Fast-DetectGPT & 0.6992 & 0.8793 & 0.7892 & 0.7961 & 0.8267 & 0.8114 & 0.8003  \\ 
    Binoculars & 0.9549 & 0.9387 & 0.9468 & 0.9240 & 0.8754 & 0.8997 & 0.9232  \\ 
    FourierGPT & 0.9380 & \textbf{0.9956} & 0.9668 & 0.8928 & 0.7976 & 0.8452 & 0.9060  \\
    RepreGuard & 0.9482 & 0.9950 & \textbf{0.9716} & 0.8882 & 0.7215 & 0.8048 & 0.8882 \\
    \midrule
    \textsc{WaveDetect-base} & 0.9249 & 0.6915 & 0.8082 & 0.8511 & 0.8382 & 0.8447 & 0.8264  \\ 
    \textsc{WaveDetect-all} & \textbf{0.9554} & 0.9619 & 0.9586 & \textbf{0.9608} & \textbf{0.9461} & \textbf{0.9535} & \textbf{0.9560}  \\  
    \bottomrule
    \end{tabular}
    }
    \caption{\textbf{OOD Generalization on Domain Shift.} We demonstrate the AUROC performance on specialized domains (Medical and Law), ensuring that these domains were not seen during training.}
    \label{tab:domainshift}
\end{table*}

When comparing \textsc{WaveDetect-all} with \textsc{WaveDetect-base}, we observe that although \textsc{WaveDetect-base} outperforms other baselines in overall performance, it shows a noticeable decline in the zero-width-pad and homoglyph tasks. This suggests that while the model trained on RAID-base performs relatively well against adversarial attacks, its robustness is still insufficient. Incorporating domain-specific data for augmentation during training can significantly enhance the model's robustness against adversarial attacks.

\wavedetect experiences its largest performance drop on the paraphrase task; however, a cross-comparison with other baselines reveals that its margin of decline is smaller. Paraphrasing manifests as changes in structural features such as wording and syntax, which demonstrates that \wavedetect captures high-level intrinsic features of MGT and is less susceptible to variations in text structure.

\subsection{RQ2: Effectiveness under LLM Evolution}

A persistent challenge in deepfake detection is ``model evolution'', as detectors often become obsolete as LLMs are updated. Table \ref{tab:evobench} investigates the stability of \wavedetect across different versions of closed-source models. \textsc{WaveDetect-base} demonstrates superior temporal stability, achieving the highest mean AUROC and the lowest performance fluctuation (Std and max$|\Delta|$) on GPT-4o, GPT-4, and Claude-Sonnet. Meanwhile, \textsc{WaveDetect-all} achieves competitive performance, very close to that of Binoculars. Mixed training with adversarial data harms the performance of \wavedetect to some degree, as it blurs the demarcation line between human texts and model texts. However, the underlying probabilistic signature can still be captured by our surrogate model and CWT, and therefore remains relatively stationary. Although Binoculars performs competitively with \wavedetect on AUROC, \wavedetect often offers more consistent performance across these model families. FourierGPT and RepreGuard perform well in stability, but poorly in AUROC. In sum, \wavedetect achieves strong overall performance in both AUROC and temporal stability.

\subsection{RQ3: Performance under Domain-Shift}

We assess the ability of detectors to generalize to OOD topics in Table \ref{tab:domainshift}.
Traditional training-based methods suffer catastrophically from domain shifts. RoBERTa and RADAR drop to near-random performances in many tasks, and even worse than random (Avg. $\sim$0.42) on average performance, likely due to overfitting on the vocabulary of the training domain.
Zero-shot detectors, Binoculars, FourierGPT and RepreGuard, achieve relatively good average performance while generalizing to other domains, especially FourierGPT and RepreGuard on medical domains. However, there exists a significant performance drop on legal domain of these methods. 
In contrast, \textsc{WaveDetect-all} exhibits remarkable robustness, achieving the highest average AUROC of \textbf{0.9560}, and performs stable across all OOD areas and all datasets. This confirms that the features captured by \wavedetect from spectral patterns of machine-generated text are ``topic-agnostic''. The rhythmic probability fluctuations exist whether the text is about clinical trials or legal precedents, irrelevant to the domain.

\section{Explanation Study}

\begin{figure}[tbp]
    \centering
    \includegraphics[width=1\linewidth]{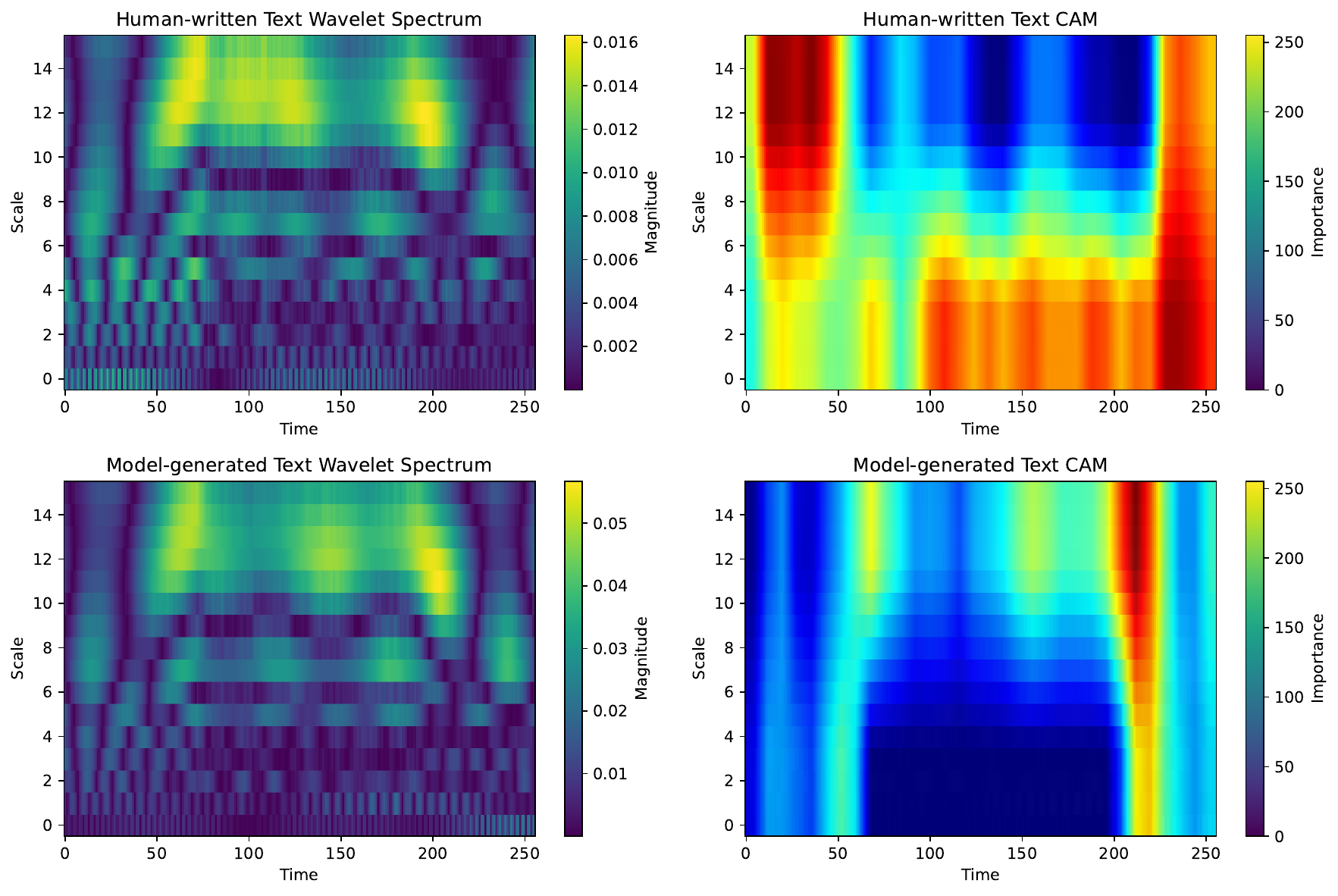}
    \caption{
    Wavelet spectrum and CAM visualizations.
    Left: wavelet spectrum representations (no clear visual separation).
    Right: CAM visualizations obtained from the trained CNN classifier applied to the wavelet spectra (highlighting discriminative regions).
    }
    \label{fig:wavelet_cam}
\end{figure}

\subsection{Spectral Mechanism Analysis}
To demystify the decision-making process of \wavedetect, we visualize both the raw wavelet spectrum and the corresponding Class Activation Maps (CAM) \cite{cam} in \Cref{fig:wavelet_cam}. This analysis reveals how the detector disentangles MGT and HWT within the time-frequency domain.

From direct visual inspection, the raw wavelet spectra of MGT and HWT do not exhibit a clear separation, as energy is broadly distributed across time-scale regions for both classes. Nevertheless, the wavelet transform remains essential by reorganizing the token-level probability sequence into a multi-scale representation that exposes scale-specific patterns beyond simplistic spectral inspection. The CAM results reveal a clear contrast in the detector's focus: activations for HWT are dominated by smaller-scale (higher-frequency) components, corresponding to short-range temporal structure, whereas MGT induces stronger responses at larger-scale (lower-frequency) regions, capturing long-range regularities. Taken together, it suggests that the CNN leverages relative, scale-dependent structures rather than absolute spectral energy, revealing discriminative cues that are otherwise obscured under direct spectrum inspection. To ensure transparency and address potential concerns regarding cherry-picking, the data provenance and additional uncurated random CAM samples are provided in Appendix \ref{app:cam_provenance}.

\subsection{Distribution of Human vs. LLM  Latent Features in \wavedetect}

\begin{figure}[tbp]
    \centering
    \includegraphics[width=\linewidth]{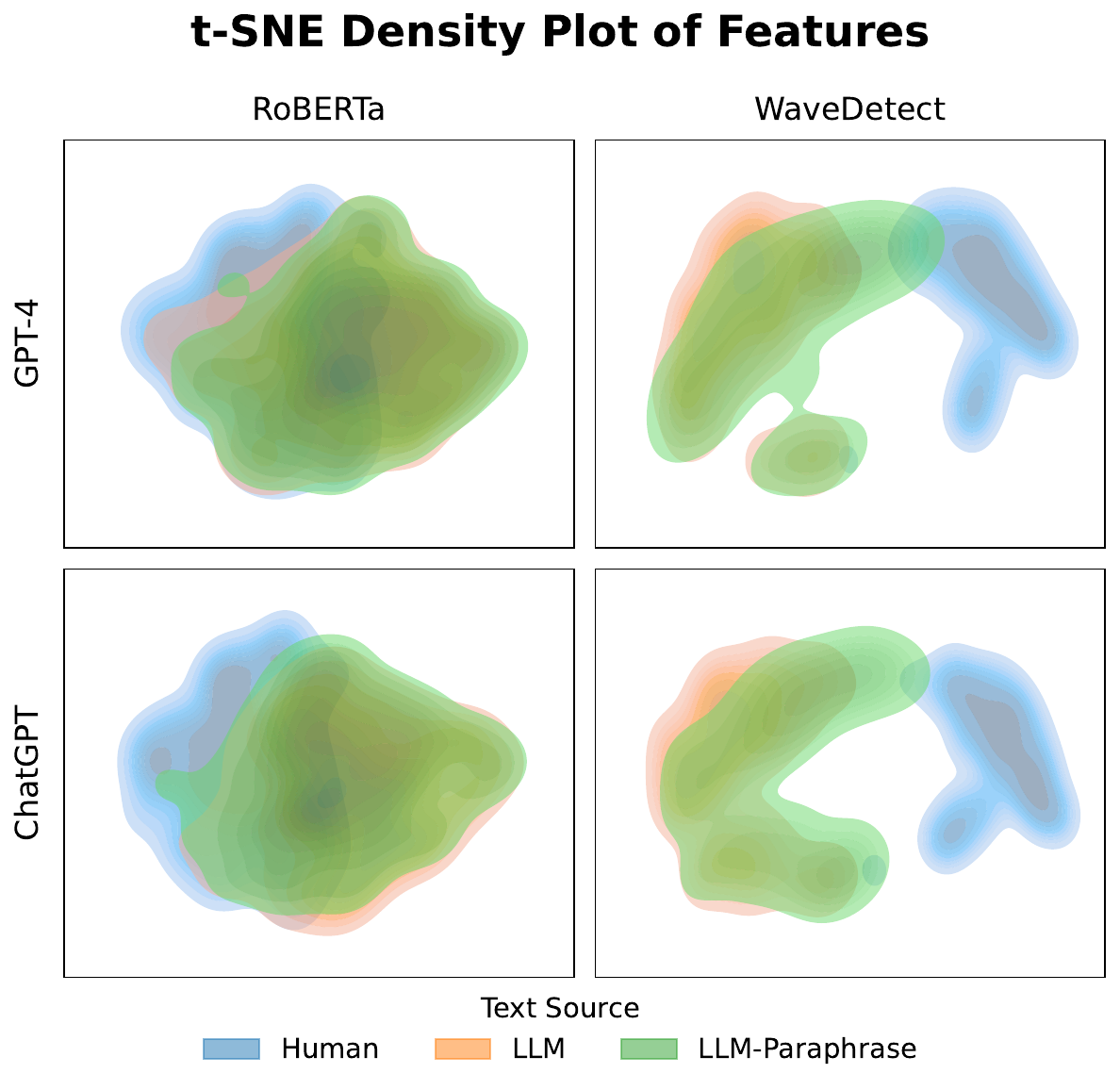}
    \caption{t-SNE visualization of dimensionality-reduced features extracted from the last layer of the detector. We present density plots representing the features of human-written text, the corresponding LLM-generated text (via GPT-4 and ChatGPT), and the LLM-generated text subjected to paraphrase attacks.}
    \label{fig:tsne}
\end{figure}

\Cref{fig:tsne} presents the t-SNE visualization of the text classification features extracted from the last layer of \wavedetect, illustrating the distributional differences between HWT and MGT (from GPT-4 and ChatGPT in RAID-test, including those subjected to paraphrase attacks). As observed in the figure, there is a clear, distinguishable gap between the two kinds. It shows that \wavedetect is able to achieve nearly 100\% accuracy in distinguishing standard GPT-4 and ChatGPT texts. However, regarding the machine texts modified by paraphrase attacks, although their distribution shape remains largely similar to the original machine texts with significant overlap, a small portion of these texts overlaps with the human text cluster. Therefore, we believe it is on these specific overlapping features that \wavedetect fails to distinguish, rendering them relatively undetectable.

\section{Conclusion}
In this paper, we introduced \wavedetect, a training-based text detection framework based on wavelet transform. Our approach converts texts to probability sequences using a surrogate model, and then captures their frequency-domain representations via wavelet transform, and utilizes a CNN to extract and aggregate the features for final classification. By jointly training the surrogate model and the CNN, we enable the model to maximally learn the discrepancies between human-written and machine-generated text. 

We carry out comprehensive evaluations on the RAID dataset, test the method's validity against adversarial attacks (RAID), LLM evolution (EvoBench), and out-of-domain tasks (Domain Shift). Experimental results indicate that \wavedetect exhibits a dominant advantage on these benchmarks, with its robustness and generalization capabilities significantly outperforming state-of-the-art baselines. These findings confirm that using spectral features can filter out noise in token-level probabilities, and thus can better capture implicit temporal dynamics of text generation. In sum, our work offers a novel and valuable perspective for further advancing the development of robust, generalizable machine-generated text (MGT) detectors.

\section{Limitations}

Our approach is currently limited to the RAID dataset, which was released over a year ago. The LLMs included in RAID are relatively old or weak compared to the current generation of models. Therefore, we believe that a detector trained on text from modern LLMs would perform even better, while adversarial attacks on modern MGTs might also prove more challenging to detect. We emphasize the urgent need for large-scale text detection datasets that incorporate modern LLMs.

Although \wavedetect exhibits strong and robust performance, the underlying mechanism of the wavelet transform requires further exploration. Specifically, factors such as the choice of mother wavelet and the range of scales need to be better understood. Our future work aims to investigate these aspects for further enhancement.

\section*{Acknowledgments}
We sincerely thank all the reviewers for their feedback on the paper. 
This study is funded by Shenzhen Science and Technology Program (No. JCYJ20240813094612017) and Guangdong Province ZJRC Program (No. 2024QN11X145).

\bibliography{custom}

\newpage
\appendix
\section{RAID Details} \label{app:raid}
The details of RAID benchmark construction are listed as follows:

\subsection{Models}
\begin{itemize}[leftmargin=*, itemsep=0em, topsep=0.5em]
    \item GPT-4 (\texttt{gpt-4-0613}), ChatGPT (\texttt{gpt-3.5-turbo-0613}), GPT-3 (text-davinci-002), GPT-2 XL \cite{gpt2, gpt3}
    \item MPT-30B, MPT-30B-Chat \cite{mpt} 
    \item Mistral-7B (Mistral-7B-v0.1), Mistral-7B-Chat (Mistral-7B-Instruct-v0.1) \cite{mistral}
    \item Cohere, Cohere-Chat \cite{cohere}
    \item Llama2-70B-Chat \cite{touvron2023llama2openfoundation}
\end{itemize}

\subsection{Dataset} %Here list the dataset used for MGTs generation.

\begin{table}[htbp]
\centering
\resizebox{\linewidth}{!}{
\begin{tabular}{l|c|r}
\toprule
\textbf{Dataset} & \textbf{Genre} & \textbf{Size} \\
\midrule
\cite{paul2021arxivabstracts}      & Abstracts & 1966 \\
\cite{bamman2013alignment}    & Books     & 1981 \\
\cite{raychev2016probabilistic}     & Code      & 920  \\
\cite{greene2006diagonal}& News      & 1980 \\
\cite{arman2020poems}                & Poetry    & 1971 \\
\cite{bien2020recipenlg}          & Recipes   & 1972 \\
\cite{volske2017tldr}        & Reddit    & 1979 \\
\cite{maas2011learning}          & Reviews   & 1143 \\
\cite{bhat2023gptwiki}         & Wiki      & 1979 \\
\cite{bohacek2022czechnews}       & Czech     & 1965 \\
\cite{schabus2017onemillion}       & German    & 1970 \\
\bottomrule
\end{tabular}
}
\caption{Details of data source in RAID.}
\label{tab:raid-details}
\end{table}

% \subsection{Decoding Strategy}
% RAID contains 3 decoding settings: greedy decoding, sampling with temperature=1 and top\_p=1, sampling with temperature=1 and top\_p=1 and repetition\_penalty=1.2.

\subsection{Decoding Strategy Evaluation}

RAID contains 3 decoding settings: greedy decoding, sampling with temperature=1 and top\_p=1, sampling with temperature=1 and top\_p=1 and repetition\_penalty=1.2.

\subsection{Adversarial Attacks}

\paragraph{Spelling and Character-Level Attacks}

\begin{itemize}[leftmargin=*, itemsep=0em, topsep=0.5em]
    \item \textbf{Alternative Spelling:} Replaces American English spellings with British English variants (e.g., ``favorite'' to ``favourite'') using a predefined mapping dictionary.
    \item \textbf{Misspelling:} Inserts common human misspellings based on a manually constructed dictionary.
    \item \textbf{Homoglyph:} Swaps standard ASCII characters with non-standard Unicode characters that look identical to the human eye (e.g., replacing a Latin `e' with a Cyrillic `e')
    \item \textbf{Upper-Lower Swap:} Randomly selects a percentage of tokens and flips the case of their first letter (uppercase to lowercase or vice versa).
    \item \textbf{Zero-Width Space:} Inserts the invisible Unicode character $U+200B$ before and after every visible character in the generation.
\end{itemize}

\paragraph{Lexical and Structural Attacks}

\begin{itemize}[leftmargin=*, itemsep=0em, topsep=0.5em]
    \item \textbf{Synonym Swap:} Replaces tokens with highly similar candidates identified by BERT. Candidates are filtered by part-of-speech tags and FastText embedding cosine similarity to ensure high-quality, diverse substitutions.
    \item \textbf{Paraphrase:} Utilizes the DIPPER-11B model (a fine-tuned T5-11B) to rewrite the text while maintaining semantic meaning.
    \item \textbf{Article Deletion:} Searches for and deletes the articles ``a'', ``an'', and ``the'' at a fixed mutation rate.
    \item \textbf{Number Swap:} Identifies numerical digits using regular expressions and replaces a sampled percentage of them with a random alternate digit between 0 and 9.
    \item \textbf{Whitespace Addition:} Randomly selects inter-token spaces and adds an extra space character to simulate irregular formatting.
    \item \textbf{Insert Paragraphs:} Splits sentences and inserts double newline characters (\texttt{\textbackslash n\textbackslash n}) between a sampled percentage of them to simulate paragraph breaks.
\end{itemize}

\section{EvoBench}

\begin{table}[t!]
\centering
\resizebox{\linewidth}{!}{
\begin{tabular}{ll}
\toprule
\textbf{Evolving LLMs} & \textbf{Source} \\
\midrule
GPT-4o & gpt-4o-2024-05-13 \\
GPT-4o & gpt-4o-2024-08-06 \\
GPT-4o & gpt-4o-2024-11-20 \\
GPT-4o & chatgpt-4o-latest \\
\midrule
GPT-4o-mini & gpt-4o-mini-2024-07-18 \\
\midrule
GPT-4 & gpt-4-0613 \\
GPT-4 & gpt-4-1106-preview \\
GPT-4 & gpt-4-0125-preview \\
GPT-4 & gpt-4-turbo-2024-04-09 \\
\midrule
Claude-Sonnet & claude-3-sonnet-20240229 \\
Claude-Sonnet & claude-3-5-sonnet-20240620 \\
Claude-Sonnet & claude-3-5-sonnet-20241022 \\
\midrule
Claude-Haiku & claude-3-haiku-20240307 \\
Claude-Haiku & claude-3-5-haiku-20241022 \\
\midrule
Claude-Opus & claude-3-opus-20240229 \\
\midrule
Gemini-Flash & gemini-1.5-flash \\
Gemini-Flash & gemini-1.5-flash-exp-0827 \\
Gemini-Flash & gemini-1.5-flash-latest \\
\midrule
Qwen & Qwen/Qwen1.5-7B-Chat \\
Qwen & Qwen/Qwen2-7B-Instruct \\
Qwen & Qwen/Qwen2.5-7B-Instruct \\
\midrule
LLaMA3 & meta-llama/Meta-Llama-3.1-8B-Instruct \\
LLaMA3 & meta-llama/Meta-Llama-3.1-70B-Instruct \\
LLaMA3 & meta-llama/Meta-Llama-3.2-1B-Instruct \\
LLaMA3 & meta-llama/Meta-Llama-3.2-3B-Instruct \\
LLaMA3 & meta-llama/Meta-Llama-3.3-70B-Instruct \\
\midrule
Fine-tuning & meta-llama/Llama-2-7b-chat-hf \\
Fine-tuning & lmsys/vicuna-7b-v1.5 \\
Fine-tuning & WizardLMTeam/WizardMath-7B-V1.0 \\
\midrule
Pruning & princeton-nlp/Sheared-LLaMA-1.3B \\
Pruning & princeton-nlp/Sheared-LLaMA-1.3B-Pruned \\
Pruning & princeton-nlp/Sheared-LLaMA-2.7B \\
Pruning & princeton-nlp/Sheared-LLaMA-2.7B-Pruned \\
\bottomrule
\end{tabular}
}
\caption{Details of LLMs in EvoBench.}
\label{tab:evobench_llms}
\end{table}

EvoBench is designed to evaluate the generalization capabilities of detectors against the real-world challenge of continuously evolving LLMs, which change over time through version updates, fine-tuning, and pruning. In our experiments, we mainly focus on the challenge brought by version evolution in closed-source LLMs. 

The principle for model selection is to identify those with multiple revisions for a single scale, where no significant alterations occurred during pre-training, resulting in a count of >2 valid revisions. Therefore, we choose GPT-4o, GPT-4, Claude-Sonnet and Gemini-Flash in our experiments.

We use MGTs over these three domains in EvoBench: XSum \cite{xsum}, WritingPrompts \cite{writingprompts} and PubMed \cite{pubmedqa}.

\section{Domain Shift}
Domain Shift is a new problem proposed by \citet{divscore}. It aims to evaluate whether detectors could maintain their capabilities in specialized, high-stakes domains like medicine and law. Details about datasets used in domain shift are described in \Cref{sec:experiment setup}. For MGT curation, they select 1000 pairs of texts for each dataset, and use GPT-4o, O3-mini, DeepSeek-V3 \cite{deepseek-v3} and DeepSeek-R1 \cite{deepseek-r1} to generate machine texts.

\section{Metrics}
To describe the metrics we used, we first need to define the basic components: True Positives ($\text{TP}$), False Positives ($\text{FP}$), True Negatives ($\text{TN}$), and False Negatives ($\text{FN}$). Based on these, the True Positive Rate (TPR) and False Positive Rate (FPR) are defined as:
\begin{equation}
    \text{TPR} = \frac{\text{TP}}{\text{TP} + \text{FN}}, \quad \text{FPR} = \frac{\text{FP}}{\text{FP} + \text{TN}}
\end{equation}

\subsection{AUROC}
The Area Under the Receiver Operating Characteristic Curve (AUROC) is a widely used threshold-independent metric that assesses the classifier's ability to distinguish between classes.

The ROC curve is generated by plotting the TPR against the FPR at various threshold settings. The AUROC is calculated as the definite integral of the ROC curve:
\begin{equation}
    \text{AUROC} = \int_{0}^{1} \text{TPR}(\text{FPR}) \, d(\text{FPR})
\end{equation}
The AUROC value ranges from 0 to 1. An AUROC of 0.5 indicates random guessing, while an AUROC of 1.0 represents a perfect classifier. Intuitively, the AUROC corresponds to the probability that a randomly selected positive instance is ranked higher than a randomly selected negative instance.

\subsection{TPR@0.1\%FPR}
While AUROC provides an overall summary of performance, it may obscure weaknesses in the low false-positive region, which is critical for safety-sensitive applications (e.g., fraud detection, medical diagnosis, or biometrics).

TPR@0.1\%FPR measures the True Positive Rate specifically when the False Positive Rate is strictly controlled at $0.1\%$ (i.e., $10^{-3}$). Let $\tau$ be the decision threshold. We find $\tau^*$ such that the FPR is fixed at the target value:
\begin{align}
    \text{TPR@0.1\%FPR} = \text{TPR}&(\tau^*) \\ 
    \nonumber \text{subject to}& \ \ \text{FPR}(\tau^*) \le 0.001
\end{align}
This metric evaluates how much of the positive class can be recalled while maintaining a very low frequency of false alarms.

\section{Computational Overhead and Training Cost}
A key advantage of \wavedetect is its efficiency during inference compared to massive LLM-based detectors. The spectral encoder uses a ResNet-18 backbone consisting of only 11.7M parameters. Table \ref{tab:complexity} compares the theoretical inference complexity of our method against strong baselines. \wavedetect only requires one forward pass of a lightweight 0.5B surrogate model and a CNN, making it highly efficient for large-scale deployment.

\begin{table}[htbp]
\centering
\resizebox{0.9\linewidth}{!}{
\begin{tabular}{ll}
\toprule
\textbf{Method} & \textbf{Complexity / Inference Cost} \\
\midrule
RoBERTa & 1 fwd of 0.1B $\sim$ 0.4B \\
RADAR & 1 fwd of 7B \\
Fast-DetectGPT & 1 fwd of 2.7B $\sim$ 8B \\
Binoculars & 2 fwd of 7B \\
FourierGPT & 1 fwd of 7B \\
RepreGuard & 1 fwd of 2B $\sim$ 7B \\
\midrule
\textbf{\wavedetect (Ours)} & \textbf{1 fwd of 0.5B + CNN (11.7M)} \\
\bottomrule
\end{tabular}
}
\caption{Computational complexity and inference cost comparison across different detectors.}
\label{tab:complexity}
\end{table}

Regarding training, \wavedetect is trained on 4 NVIDIA RTX 6000 Ada GPUs for 5 hours and 33 minutes. Total training consisted of approximately 20.1k steps.

\section{Robustness under Different Decoding Strategies}
To further investigate the impact of these strategies, we break down the performance of \wavedetect and several baselines across these specific decoding settings. As shown in Table \ref{tab:decoding_breakdown}, \wavedetect maintains a strong and stable performance across all decoding variants, particularly outperforming baselines in sampling without repetition penalty.

\begin{table}[htbp]
\centering
\resizebox{\linewidth}{!}{
\begin{tabular}{lccc}
\toprule
\textbf{Method} & \textbf{Greedy} & \textbf{Sampling (RP=Yes)} & \textbf{Sampling (RP=No)} \\
\midrule
Binoculars & 0.8957 & 0.4614 & 0.8928 \\
RoBERTa & 0.6432 & 0.4385 & 0.6152 \\
Fast-DetectGPT & 0.8129 & 0.4186 & 0.8247 \\
RADAR & 0.8636 & 0.7099 & 0.8203 \\
RepreGuard & 0.7070 & 0.5185 & 0.6386 \\
\textsc{WaveDetect-Base} & 0.8780 & 0.6432 & 0.8321 \\
\textsc{WaveDetect-All} & \textbf{0.9895} & \textbf{0.9578} & \textbf{0.9712} \\
\bottomrule
\end{tabular}
}
\caption{Performance breakdown across different decoding strategies on the RAID dataset. RP denotes Repetition Penalty.}
\label{tab:decoding_breakdown}
\end{table}

\section{Ablation Study: CWT vs. STFT}

We compare the Continuous Wavelet Transform (CWT) with a Short-Time Fourier Transform (STFT) baseline for modeling token probability dynamics. Both models are trained on the same dataset to ensure a fair comparison.

\subsection{STFT Baseline}
We apply STFT with a Hann window, using \texttt{win\_length=256}, \texttt{n\_fft=256}, and \texttt{hop\_length=64}. Short sequences are zero-padded. Since STFT reduces temporal resolution, we use bilinear interpolation to restore the sequence length $\text{T}$.

\subsection{Results}
Table~\ref{tab:ablation_results} reports mean AUROC across different evaluation settings. Specifically, \textbf{RAID} denotes the average AUROC over the full benchmark; \textbf{Evobench} reports the average AUROC across four evolving LLM families (GPT-4o, GPT-4, Claude-3-Sonnet, and Gemini-1.5-Flash); and \textbf{DivScore} is the average AUROC over 4$\times$4 settings, including four models (DeepSeek-R1, DeepSeek-V3, GPT-4o, O3-mini) and four datasets (LawStack, OALC, MIMIC, PubMed).

\begin{table}[htbp]
\centering
\resizebox{0.9\linewidth}{!}{
\begin{tabular}{lccc}
\toprule
\textbf{Method} & \textbf{RAID} & \textbf{DivScore} & \textbf{Evobench} \\
\midrule
STFT            & 0.9570 & 0.9000 & 0.7904 \\
CWT         & \textbf{0.9740} {\scriptsize(+1.70\%)} 
                & \textbf{0.9500} {\scriptsize(+5.00\%)} 
                & \textbf{0.7948} {\scriptsize(+0.44\%)} \\
\bottomrule
\end{tabular}
}
\caption{Performance comparison (mean AUROC) and gains of CWT over STFT.}
\label{tab:ablation_results}
\end{table}

\subsection{Discussion}
CWT outperforms STFT due to its multi-resolution nature. Unlike STFT's fixed window (limited by the time-frequency trade-off), CWT captures both local spikes and global trends in non-stationary token probability signals, leading to more robust and generalizable features.

\section{Visualization Provenance and Additional Samples} \label{app:cam_provenance}
To ensure transparency and address potential concerns regarding cherry-picking raised during the review process, we clarify the provenance of our visual analytics. The CAM visualizations in \Cref{fig:wavelet_cam} and the t-SNE density plots in \Cref{fig:tsne} of the main text were generated using samples explicitly drawn from the \textbf{RAID test set}. Specifically, we utilized standard GPT-4 and ChatGPT outputs, alongside their adversarially paraphrased counterparts, to illustrate the distributional shifts.

\begin{figure*}[htbp] 
    \centering
    
    \begin{subfigure}{0.48\textwidth}
        \centering
        \includegraphics[width=\linewidth]{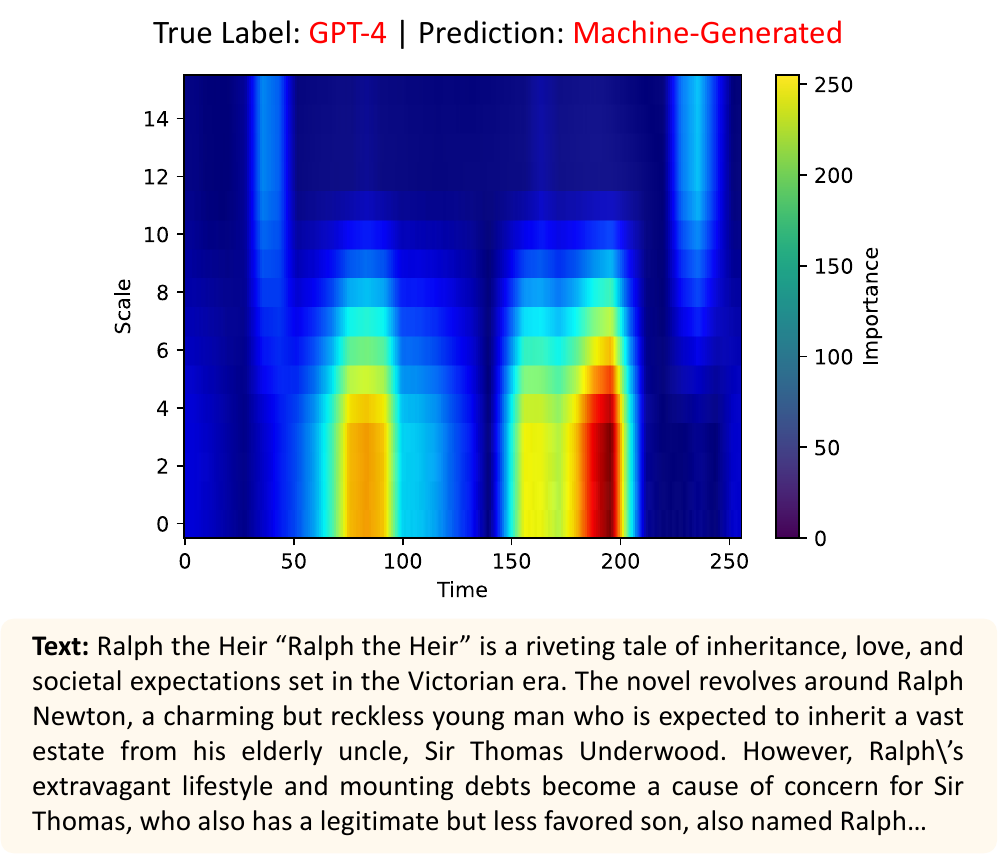}
        \caption{Machine-Generated Sample 1}
        \label{fig:cam_mgt1}
    \end{subfigure}
    \hfill
    \begin{subfigure}{0.48\textwidth}
        \centering
        \includegraphics[width=\linewidth]{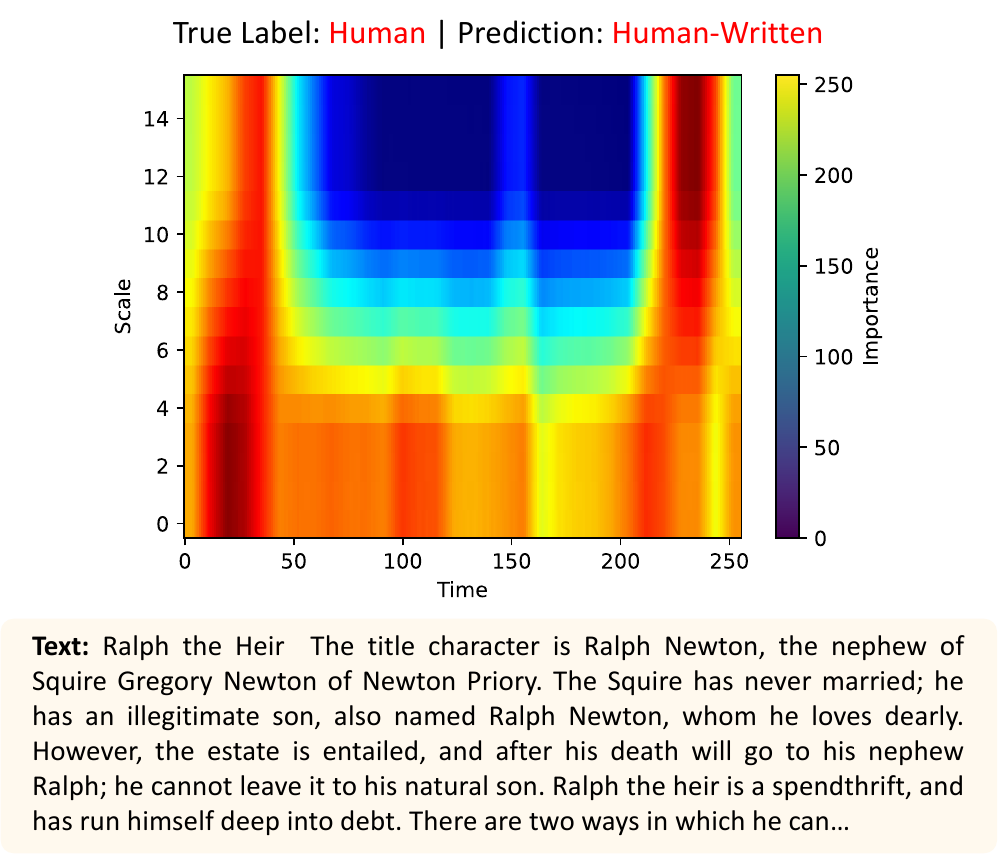}
        \caption{Human-Written Sample 1}
        \label{fig:cam_hum1}
    \end{subfigure}
    
    \vspace{0.3cm}
    
    \begin{subfigure}{0.48\textwidth}
        \centering
        \includegraphics[width=\linewidth]{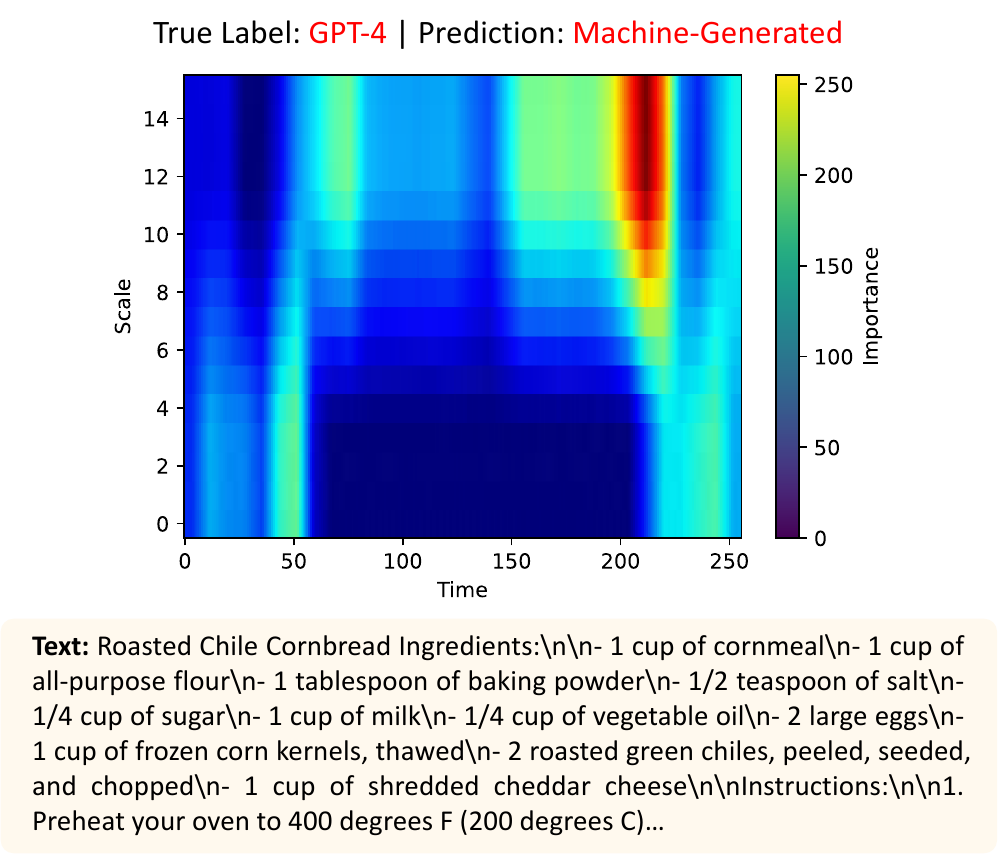}
        \caption{Machine-Generated Sample 2}
        \label{fig:cam_mgt2}
    \end{subfigure}
    \hfill
    \begin{subfigure}{0.48\textwidth}
        \centering
        \includegraphics[width=\linewidth]{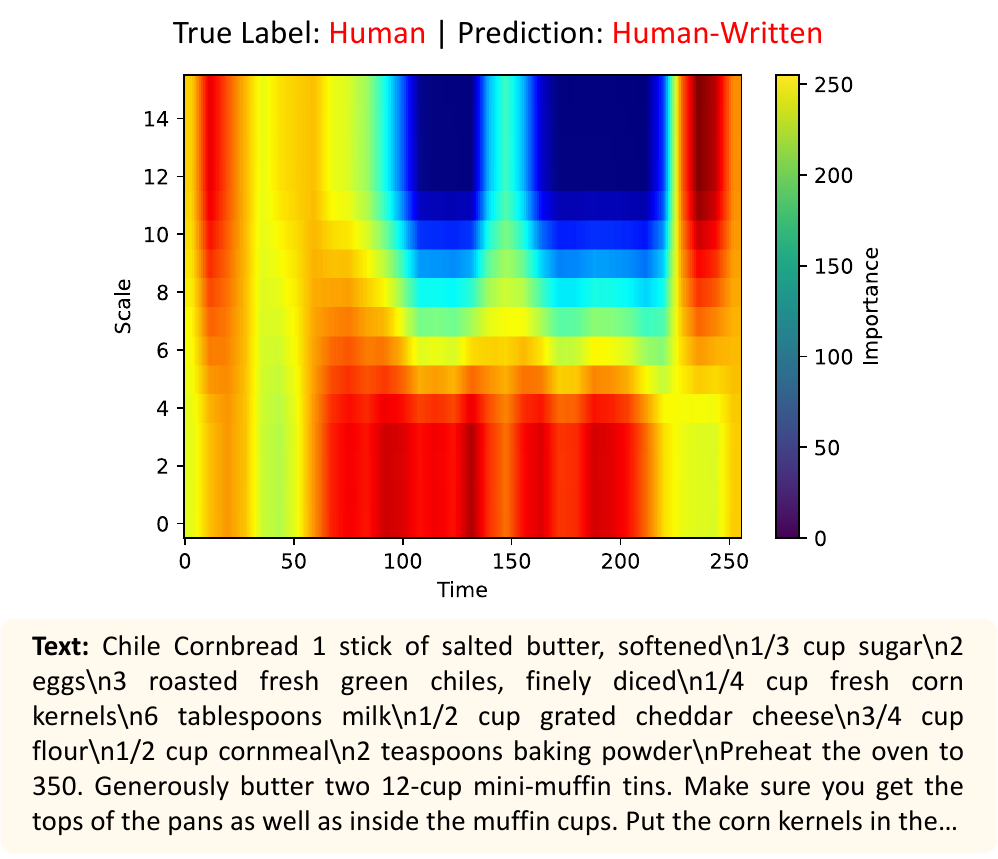}
        \caption{Human-Written Sample 2}
        \label{fig:cam_hum2}
    \end{subfigure}
    
    \vspace{0.3cm}
    
    \begin{subfigure}{0.48\textwidth}
        \centering
        \includegraphics[width=\linewidth]{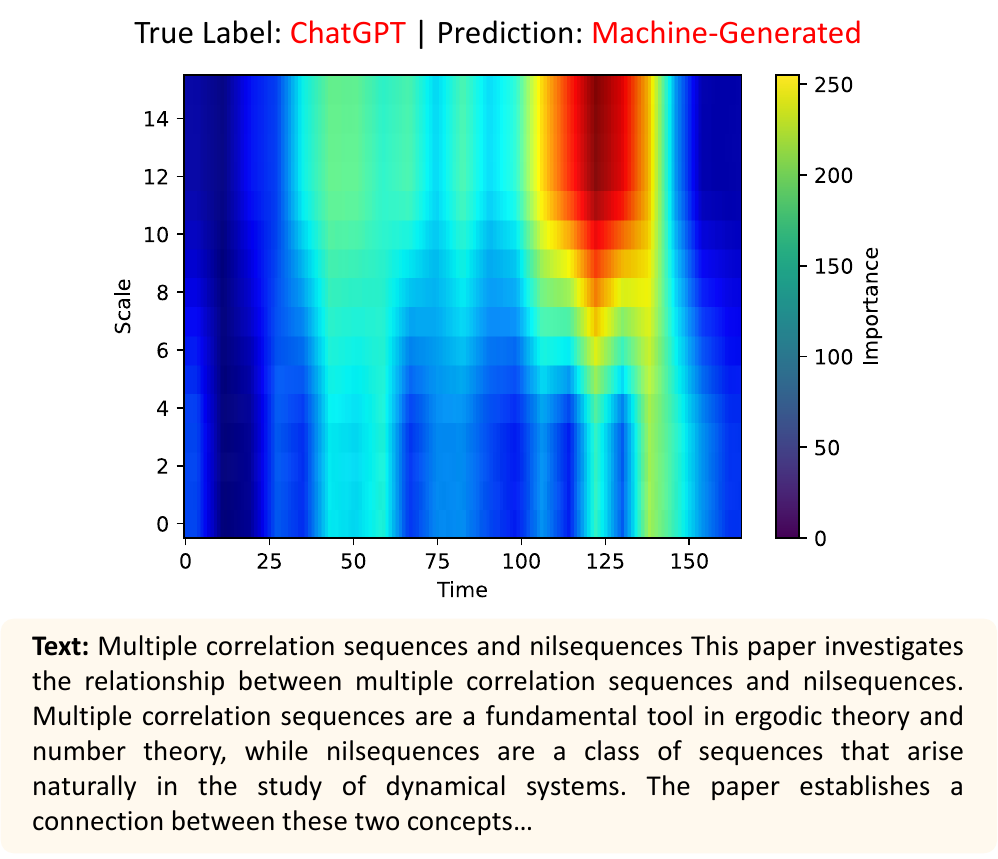}
        \caption{Machine-Generated Sample 3}
        \label{fig:cam_mgt3}
    \end{subfigure}
    \hfill
    \begin{subfigure}{0.48\textwidth}
        \centering
        \includegraphics[width=\linewidth]{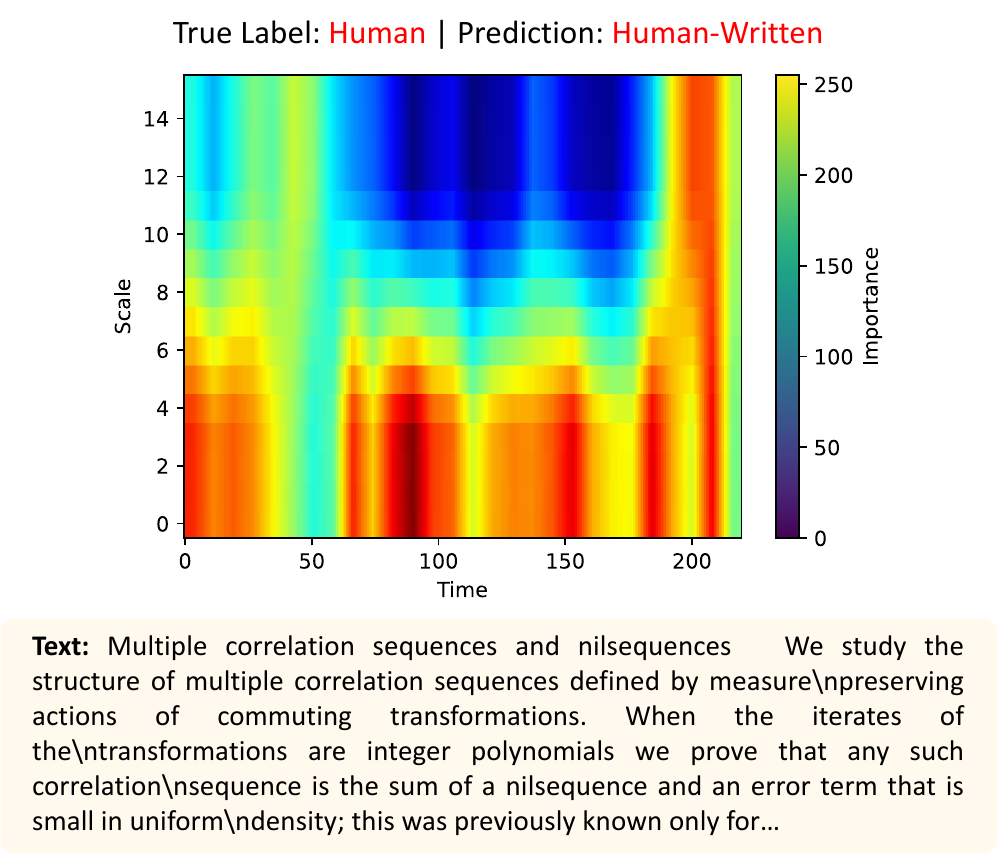}
        \caption{Human-Written Sample 3}
        \label{fig:cam_hum3}
    \end{subfigure}

    \caption{Additional uncurated random samples of Wavelet spectrums and CAM visualizations drawn from the RAID test set. The left column (\ref{fig:cam_mgt1}, \ref{fig:cam_mgt2}, \ref{fig:cam_mgt3}) displays machine-generated texts, characterized by dominant large-scale (low-frequency) activations. The right column (\ref{fig:cam_hum1}, \ref{fig:cam_hum2}, \ref{fig:cam_hum3}) displays human-written texts, characterized by scattered, small-scale (high-frequency) activations.}
    \label{fig:random_cams}
\end{figure*}

In Figure \ref{fig:random_cams}, we provide additional CAM visualizations of randomly sampled texts from both the human and LLM. These non-curated, random examples further demonstrate that the discriminative spectral patterns captured by our CNN--specifically, the high-frequency transients in HWTs and low-frequency regularities in MGTs--are consistent and statistically robust across the broader dataset.

\end{document}